\documentclass[preprint,3p,times,twocolumn]{elsarticle}

\usepackage{amssymb}
\usepackage{amsthm}

\usepackage{lineno}

\usepackage{amsmath, amsfonts}
\usepackage{mathabx}   
\usepackage[makeroom]{cancel}    
\usepackage{tipa}      
\usepackage{textcomp}  
\usepackage{latexsym}  
\usepackage{mathrsfs}  

\usepackage{graphicx}
\usepackage{tikz}
\usepackage{pgfplots}
\pgfplotsset{compat=1.6}
\usepgfplotslibrary{groupplots}
\usepackage{pgf}
\usetikzlibrary{positioning,shapes,arrows,fit,calc,automata,perspective,3d}
\usepackage{neuralnetwork}
\usepackage{tikz-network}

\usepackage{hyperref}
\hypersetup{
	colorlinks=true,       
	linkcolor=blue,
	citecolor=red,
	filecolor=magenta,      
	urlcolor=cyan           
}
\usepackage{cleveref}

\usepackage{hhline}

\usepackage{multirow}
\usepackage{colortbl} 

\usepackage{algpseudocode}
\usepackage{algorithm}

\usepackage{subcaption} 

\newtheorem{defin}{Definition}[section]%

\theoremstyle{defin}
\newtheorem{rem}{Remark}[section]

\newcommand{\red}[1]{#1}

\newcommand{\TENS}[1]{\mathop{\mathbf{#1}}\nolimits} 
\newcommand{\norm}[1]{\Vert{#1}\Vert}	
\newcommand{\Id}[1]{\ensuremath{\mathbb{I}_{#1}}}   
\newcommand{\pe}[1]{^{(#1)}}	

\newcommand{\N}{\ensuremath{\mathbb{N}}}    
\newcommand{\R}{\ensuremath{\mathbb{R}}}	

\DeclareMathOperator{\diag}{diag}

\newcommand{\wh}{\widehat}	

\renewcommand{\v}{\ensuremath{\boldsymbol}}	

\journal{arXiv}
\date{}

\begin{document}

\begin{frontmatter}



\title{Edge-Wise Graph-Instructed Neural Networks}


\author[inst1,inst3]{Francesco Della Santa\corref{cor1}}
\cortext[cor1]{Corresponding author}

\affiliation[inst1]{organization={Department of Mathematical Sciences, Politecnico di Torino},
            addressline={Corso Duca degli Abruzzi 24}, 
            postcode={10129}, 
            state={Turin},
            country={Italy}}

\affiliation[inst3]{organization={Gruppo Nazionale per il Calcolo Scientifico INdAM},
            addressline={Piazzale Aldo Moro 5}, 
            postcode={00185}, 
            state={Rome},
            country={Italy}}

\author[inst2]{Antonio Mastropietro}
\author[inst1,inst3]{Sandra Pieraccini}
\author[inst1]{Francesco Vaccarino}

\affiliation[inst2]{organization={Department of Computer Science, University of Pisa},
            addressline={Largo B. Pontecorvo 3}, 
            postcode={56127}, 
            state={Pisa},
            country={Italy}}

\begin{abstract}

The problem of multi-task regression over graph nodes has been recently approached through
Graph-Instructed Neural Network (GINN), which is a promising architecture belonging to the subset of message-passing graph neural networks.
In this work, we discuss the limitations of the Graph-Instructed (GI) layer, and we formalize a novel edge-wise GI (EWGI) layer. 
We discuss the advantages of the EWGI layer and we provide numerical evidence that EWGINNs perform better than GINNs over \red{some graph-structured input data, like the ones inferred from the Barab\'asi-Albert graph, and improve the training regularization on graphs} with chaotic connectivity, like the ones inferred from the Erdos-R\'enyi graph.

\end{abstract}





\begin{keyword}
Graph Neural Networks\sep Deep Learning\sep Regression on Graphs
\MSC[2020] 05C21\sep 
65D15\sep 
68T07\sep 
90C35 
\end{keyword}

\end{frontmatter}

\section{Introduction}\label{sec:intro}

Graph Neural Networks (GNNs) are powerful tools for learning tasks on graph-structured data \cite{GNNsurvey2020}, such as node classification \cite{maurya2022simplifying}, link prediction, or graph classification. Their formulation traces back to the late 2000s \cite{firstGNN_Gori2005,firstGNN_Micheli2009,firstGNN_Scarselli2009}. 
In the last years, GNNs have received increasing attention from the research community for their application in biology \cite{cinaglia2024multilayer}, chemistry \cite{atz2021geometric, gilmer2017neural}, finance \cite{cheng2022financial}, geoscience \cite{GINN}, computational social science \cite{aref2024analyzing}, and particle physics \cite{dezoort2023graph}, to name a few.
\red{Among the available models in the literature, we mention Graph ConvNet, GraphSage, and Graph Attention Networks as models for tasks such as graph, node, or edge classification, or for graph regression \cite{dwivedi2023benchmarking, dwivedi2022long}.}
Yet, the community has neglected the applications concerning the Regression on Graph Nodes (RoGN) learning task. 
Indeed, to the best of the authors' knowledge, the most used benchmarks do not include datasets for this task \cite{dwivedi2023benchmarking, dwivedi2022long}. 
\red{Nonetheless, there is an increasing interest in RoGN, especially among researchers working on physics-based simulations where, for example, predictions on mesh or grid nodes are performed (see for example \cite{PICHI2024112762}).

RoGN can be stated as multi-task regression, where the input data are endowed with a graph structure.}
The benchmark models for multi-task regression are Fully Connected Neural Networks (FCNNs).
Recently, a new type of layer for GNNs has been developed in \cite{GINN}, belonging to the class of message-passing GNNs \cite{gilmer2017neural}.
From now on we will refer to these layers as Graph-Instructed (GI) layers; 
Graph-Instructed NNs (GINNs) are built by stacking GI layers. 
GINNs have demonstrated good performance on RoGN, showing better results than FCNNs, as illustrated in \cite{GINN}.
Although the GINN architecture has been specifically designed for RoGN, the usage of GI layers has been recently extended to supervised classification tasks (see \cite{dellasanta2024graphinformed}).

We point the reader to the fact that in \cite{GINN} GI layers and GINNs are denoted as Graph-Informed layers and Graph-Informed NNs, respectively.
In \cite{HALL2021110192}, in a different framework from the one addressed in \cite{GINN}, a homonymous but different model is presented; therefore, to avoid confusion with \cite{HALL2021110192}, we have changed the names of both layers and NNs.

GI layers are based on a weight-sharing principle, such that their weights rescale the outgoing message from each node.
In this paper, to improve the generalization capability of their inner-layer representation,  we introduce Edge-Wise Graph-Instructed (EWGI) layers, characterized by additional weights (associated with graph nodes) that enable the edge-wise customization of the passage of information to each receiving node.

We compare the Edge-Wise GINN (EWGINN) with the GINN in the experimental settings originally used in \cite{GINN} for validating the models; these settings are RoGN tasks on two stochastic flow networks based on a Barab\'asi-Albert graph and an Erdos-R\'enyi graph, respectively.
In particular, we show that EWGINNs perform better on the Barab\'asi-Albert connectivity structure, with a small increment of the number of learning weights.

The work is organized as follows: in \Cref{sec:GIlayers} the GI layers are introduced, recalling their inner mechanisms. 
\Cref{sec:ewgi_layers} formally defines EWGI layers and theoretically discusses their properties.
Then, in \Cref{sec:experiments}, we analyze the experiment results for the RoGN tasks, comparing with the previous literature \cite{GINN}.
Finally, \Cref{sec:conclusions} summarizes our work and discusses future improvements and research directions.

\section{Graph-Instructed Layers}\label{sec:GIlayers}
This section briefly reviews previous GINNs to establish the framework for introducing our main contribution.
Graph-Instructed (GI) Layers are NN layers defined by an alternative graph-convolution operation introduced in \cite{GINN}. Given a graph $G$ (without self-loops) and its adjacency matrix $A\in\R^{n\times n}$, a basic GI layer for $G$ is a NN layer with one input feature per node and one output feature per node described by a function $\mathcal{L}^{GI}:\R^n\rightarrow\R^n$ such that
\begin{equation}\label{eq:GI_action_simple}
    \begin{aligned}
    \mathcal{L}^{GI}(\v{x}) 
    = 
    \v{\sigma}\left( (\diag(\v{w}) (A + \Id{n}))^T\, \v{x} + \v{b}\right),
    \end{aligned}
    \,
\end{equation}
for each vector of input features $\v{x}\in\R^n$ and where:
\begin{itemize}
    \item $\v{w}\in\R^n$ is the weight vector, with the component $w_i$  associated to the graph node $v_i$, $i=1,\ldots , n$.

    \item $\diag(\v{w})\in\R^{n\times n}$ is the diagonal matrix with elements of $\v{w}$ on the diagonal and $\Id{n}\in\R^{n\times n}$ is the identity matrix. For future reference, we set $\wh{W}:=\diag(\v{w}) (A + \Id{n})$;

    \item $\v{\sigma}:\R^n\rightarrow\R^n$ is the element-wise application of the activation function $\sigma$;
    
    \item $\v{b}\in\R^n$ is the bias vector.
\end{itemize}

In brief, Eq. \eqref{eq:GI_action_simple} is equivalent to the action of a Fully-Connected (FC) layer where the weights are the same if the connection is outgoing from the same unit, whereas it is zero if two units correspond to graph nodes that are not connected; more precisely:
\begin{equation*}\label{eq:GI_weights_simple}
    \wh{w}_{ij}=
    \begin{cases}
    w_i\,,\quad & \text{if }a_{ij}\neq 0 \text{ or }i=j\\
    0\,,\quad & \text{otherwise}
    \end{cases}
    \,,
\end{equation*}
where $a_{ij},\wh{w}_{ij}$ denote the $(i,j)$-th element of $A, \wh{W}$, respectively.

\noindent On the other hand, from a message-passing point of view, the operation described in \eqref{eq:GI_action_simple} is equivalent to having each node $v_i$ of $G$ sending to its neighbors a message equal to the input feature $x_i$, scaled by the weight $w_i$; then, each node sum up all the messages received from the neighbors, add the bias, and applies the activation function. In a nutshell, the message-passing interpretation can be summarized by the following node-wise equation:
\begin{equation}\label{eq:ginn_node_action}
x_{i}' = \sum_{j \in \mathrm{N}_{\text{in}}(i)\cup \{i\}} x_j \, w_j  + b_i\,,
\end{equation}
where $x_i'$ is the output feature of the GI layer corresponding to node $v_i$ and $\mathrm{N}_{\rm in}(i)$ is the set of indices such that $j\in\mathrm{N}_{\rm in}(i)$ if and only if $e_{ij}=\{v_i,v_j\}$ is an edge of the graph. We dropped the action of the activation function $\sigma$ for simplicity.

Layers characterized by \eqref{eq:GI_action_simple} can be generalized to read any arbitrary number $K\geq 1$ of input features per node and to return any arbitrary number $F\geq 1$ of output features per node. Then, the general definition of a GI layer is as follows.

\begin{defin}[GI Layer - General form \cite{GINN}]\label{def:GIlayer_general}
    A \emph{GI layer with $K\in\N$ input features and $F\in\N$ output features} is a NN layer with $n F$ units connected to a layer with outputs in $\R^{n\times K}$ and having a characterizing function $\mathcal{L}^{GI}:\R^{n\times K}\rightarrow\R^{n\times F}$ defined by
    \begin{equation}\label{eq:GIlayer_general}
    	\mathcal{L}^{GI}(X) = \v{\sigma}\left( \widetilde{\TENS{W}}^T \mathrm{vertcat}(X) + B \right)\,,
    \end{equation}
    where:
    \begin{itemize}
        \item $X\in\R^{n\times K}$ is the input matrix (i.e., the output of the previous layer) and $\mathrm{vertcat}(X)$ denotes the vector in $\R^{nK}$ obtained concatenating the columns of $X$;
        
        \item tensor $\widetilde{\TENS{W}}\in \R^{nK\times F\times n}$ is the concatenation along the $2^{\rm nd}$ dimension (i.e., the column-dimension) of the matrices $\widetilde{W}^{(1)},\ldots ,\widetilde{W}^{(F)}$, defined as
        \begin{equation}\label{eq:Wtilde_filter_concat}
        	\widetilde{W}^{(l)} := 
        	\begin{bmatrix}
        		\widehat{W}\pe{1,l}\\
        		\vdots\\
        		\widehat{W}\pe{K, l}
        	\end{bmatrix} 
        	=
        	\begin{bmatrix}
        		\mathrm{diag}(\v{w}\pe{1, l})\wh{A}\\
        		\vdots\\
        		\mathrm{diag}(\v{w}\pe{K, l})\wh{A}
        	\end{bmatrix}
        	\in\R^{nK\times n}\,,
        \end{equation}
        for each $l=1,\ldots ,F$, after being reshaped as tensors in $\R^{nK\times 1\times n}$. Vector $\v{w}\pe{k,l}\in\R^n$ is the weight vector characterizing the contribution of the $k$-th input feature to the computation of the $l$-th output feature of the nodes, for each $k=1,\ldots ,K$, and $l=1,\ldots ,F$; matrix $\wh{A}$ denotes $A+\mathbb{I}_n$. 
        
        \item the operation $\widetilde{\TENS{W}}^T \mathrm{vertcat}(X)$ is a tensor-vector product and $B\in\R^{n\times F}$ is the matrix of the biases.
    \end{itemize}
\end{defin}

\noindent Additionally, pooling and mask operations can be added to GI layers (see \cite{GINN} for more details). 

From now on, we call \emph{Graph-Instructed Neural Network} (GINN) a NN made of GI layers \cite{GINN}.
We point out that the number of weights of a GI layer is equal to $nKF+nF$. On the other hand, the number of weights of a FC layer of $n$ units, reading the outputs of a layer of $m$ units, is equal to $mn + n$; therefore, if we consider the case of $m=n$ and $KF + F < n+1$ (typically satisfied for sufficiently large graphs), GI layers have fewer weights to be trained compared with the FC layer. 
Moreover, we observe that adjacency matrices are typically sparse and, therefore, the tensor $\widetilde{\TENS{W}}$ in \eqref{eq:GIlayer_general} is typically sparse too. Then, it is possible to exploit the sparsity of this tensor to reduce the memory cost of the GINN implementation.

\section{Edge-Wise Graph-Instructed Layers}\label{sec:ewgi_layers}

A possible drawback of GI layers is that their weights rescale only the outgoing information of the nodes. For example, if nodes $v_j$ and $v_k$ are connected to node $v_i$ in a graph $G=(V, E)$ such that $(v_i, v_j), (v_i, v_k)\in E$, then the units corresponding to $v_j$ and $v_k$ in a GI layer based on $G$ receive the same contribution from the input features corresponding to node $v_i$; moreover, if nodes $v_j,v_k$ have the same neighbors, the GI layer's outputs corresponding to these nodes are the same except for the contribution of the bias and the contribution from themselves.
This property is useful to reduce the number of weights per layer and, depending on the complexity of the target function defined on the graph nodes, it is not necessarily a limitation. 
Nonetheless, it surely limits the representational capacity of the model. Therefore, some target functions can be too complicated to be modeled by GI layers.

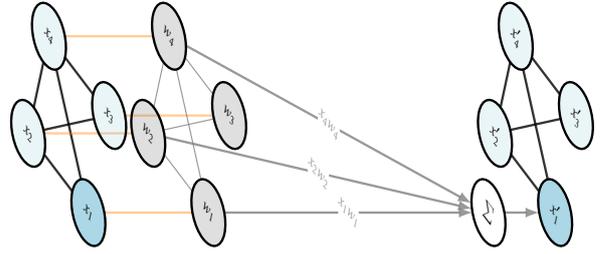
\begin{figure}[htb]
    \centering
    \resizebox{0.48\textwidth}{!}{
    \begin{tikzpicture}[multilayer=3d,rotate=90]
    \Vertices{ewginnfilter_verts_v2.csv}
    \Edges{ewginnfilter_edges_v2.csv}
    \end{tikzpicture}
	}
    \caption{Visual representation of \eqref{eq:ewginn_node_action}. Example with $n=4$ nodes (non-directed graph), $i=1$; for simplicity, the bias is not illustrated.}
    \label{fig:ewginnfilter}
\end{figure}

Given the observation above, it is useful to define a new GI layer capable of improving the capacity of the model at a reduced cost in terms of the total number of trainable weights. 
In this work, we propose to modify the classic GI layers by adding an extra set of weights associated with the nodes to rescale their incoming information. 
In brief, given the node-wise equation \eqref{eq:GI_action_simple}, we change it into
\begin{equation}\label{eq:ewginn_node_action}
x_{i}' = w_i^{\rm in}\sum_{j \in \mathrm{N}_{\text{in}}(i)\cup \{i\}} x_j \, w_j^{\rm out}  + b_i\,,
\end{equation}
where $w_{j}^{\rm out}$ denotes the (old) weights for rescaling the outgoing information from node $v_j$, while $w_{i}^{\rm in}$ denotes the (new) weights for rescaling the incoming information to node $v_i$ (see \Cref{fig:ewginnfilter}).

A NN layer based on \eqref{eq:ewginn_node_action} is a layer with one input feature per node and one output feature per node, described by a function $\mathcal{L}:\R^n\rightarrow\R^n$ such that
\begin{equation}\label{eq:ewGI_action_simple}
    \begin{aligned}
    &\mathcal{L}(\v{x})
    = 
    \v{\sigma}\left( (\diag(\v{w}^{\rm out}) (A + \Id{n}) \diag(\v{w}^{\rm in}))^T\, \v{x} + \v{b}\right)
    \end{aligned}
    \,
\end{equation}
for each vector of input features $\v{x}\in\R^n$ and where $\v{w}^{\rm out}, \v{w}^{\rm in}\in\R^n$ are the weight vectors, where the components $w_i^{\rm out}, w_i^{\rm in}$ are  associated to the graph nodes $v_i$, for each $i=1,\ldots, n$. For future reference, from now on, we set $\wh{W}:=\diag(\v{w}^{\rm out}) (A + \Id{n}) \diag(\v{w}^{\rm in})$.

In brief, \eqref{eq:ewGI_action_simple} is equivalent to a FC layer where the weights are zero if two distinct units correspond to graph nodes that are not connected, otherwise $\wh{w}_{ij} = w_i^{\rm out}w_j^{\rm in}$ if $e_{ij}\in E$ or $i=j$. Therefore, we observe that each weight $\wh{w}_{ij} = w_i^{\rm out}w_j^{\rm in}$ is associated with the edge $e_{ij}=(v_i,v_j)$ in the graph or the self-loop added by the layer (if $i=j$). 
Given the above observations, we can interpret \eqref{eq:ewGI_action_simple} as the operation of a NN layer with weights associated with edges instead of nodes.
Then, we define the new layer as \emph{Edge-Wise GI} (EWGI) Layer.

\begin{rem}[EWGI Layers - Advantages of the Formulation]\label{rem:paramsharing_ewgi}
    Note that in principle EWGI layers could be defined by associating an independent weight $\wh{w}_{ij}$ to each edge of $G$ and each added self-loops. Nonetheless, the approach here proposed exhibits the following advantages:
    \begin{itemize}
        \item If $G$ is a directed graph, we have that $n-1 < |E| < n^2-n$; therefore, for the independent weight formulation the total number of weights is in the range $[2n -1, n^2]$ (biases excluded). 
        On the other hand, in \eqref{eq:ewGI_action_simple} the number of weights is always equal to $2n$ (biases excluded).
        \item If $G$ is an undirected graph, we have $n-1 < |E| < (n^2-n)/2$; therefore, for the independent weight formulation the total number of weights is in the range $[2n-1, n+(n^2-n)/2]$ (biases excluded).
        On the other hand, in \eqref{eq:ewGI_action_simple} the number of weights is always equal to $2n$ (biases excluded).
    \end{itemize}
    The advantage of using formulation \eqref{eq:ewGI_action_simple} is therefore evident: independently of the number of graph edges, the number of weights is always $2n$, which is essentially the lower bound of the number of weights in the other formulation.
\end{rem}

Analogously to classic GI layers, EWGI layers can be generalized to read any arbitrary number $K\geq 1$ of input features per node and to return any arbitrary number $F\geq 1$ of output features per node. Then, the general definition of a EWGI layer is as follows.

\begin{defin}[EWGI Layer - General form]\label{def:ewGIlayer_general}
    An \emph{EWGI layer with $K\in\N$ input features and $F\in\N$ output features} is a NN layer with $n F$ units connected to a layer with outputs in $\R^{n\times K}$ and having a characterizing function $\mathcal{L}^{EWGI}:\R^{n\times K}\rightarrow\R^{n\times F}$ defined by
    \begin{equation}\label{eq:GIlayer_general}
    	\mathcal{L}^{EWGI}(X) = \v{\sigma}\left( \widetilde{\TENS{W}}^T \mathrm{vertcat}(X) + B \right)\,,
    \end{equation}
    where the tensor $\widetilde{\TENS{W}}\in \R^{nK\times F\times n}$ is defined as the concatenation along the $2^{\rm nd}$ dimension of the matrices $\widetilde{W}^{(1)},\ldots ,\widetilde{W}^{(F)}$, such that
        \begin{equation}\label{eq:Wtilde_filter_concat}
        	\widetilde{W}^{(l)} :=
        	\begin{bmatrix}
        		\mathrm{diag}(\v{w}\pe{1, l}_{\rm out}) \, \wh{A} \, \mathrm{diag}(\v{w}\pe{1,l}_{\rm in})\\
        		\vdots\\
        		\mathrm{diag}(\v{w}\pe{K, l}_{\rm out}) \, \wh{A} \, \mathrm{diag}(\v{w}\pe{K,l}_{\rm in})
        	\end{bmatrix}
        	\in\R^{nK\times n}\,,
        \end{equation}
        for each $l=1,\ldots ,F$, after being reshaped as tensors in $\R^{nK\times 1\times n}$, and where:
        \begin{itemize}
            \item $\v{w}\pe{k,l}_{\rm out}\in\R^n$ is the weight vector characterizing the contribution of the $k$-th input feature to the computation of the $l$-th output feature of the nodes, for each $k=1,\ldots ,K$, and $l=1,\ldots ,F$, with respect to the outgoing information;
            
            \item $\v{w}\pe{k,l}_{\rm in}\in\R^n$ is the weight vector characterizing the contribute of the $k$-th input feature to the computation of the $l$-th output feature of the nodes, for each $k=1,\ldots ,K$, and $l=1,\ldots ,F$, with respect to the incoming message.
        \end{itemize}
\end{defin}

From the definition above, we observe that the number of weights of a general EWGI layer is $2nKF+nF$. 
Therefore, if we consider a FC layer of $n$ units, reading the outputs of a layer of $m=n$ units, the EWGI layers have a smaller number of weights to be trained if $2KF + F < n+1$. 

From now on, we call \emph{Edge-Wise Graph-Instructed Neural Network} (EWGINN) a NN made of EWGI layers.

\section{Preliminary Results}\label{sec:experiments}

In this section, we illustrate the results of a preliminary experimental study about the representational capacity of the new EWGI layers and EWGINNs. 
We compare the performances of a set of EWGINNs with the \red{ones of a set of GINNs for the RoGN task of the two stochastic maximum flow problems reported in \cite{GINN}. 
In particular, we train the models using the same architectures, hyperparameters, and training options; for the EWGINNs, we replace GI layers with EWGI layers. In order to strengthen the study, we train each configuration with respect to five different random seeds, reporting the median performances for each configuration.}

\subsection{Maximum Flow Regression for Stochastic Flow Networks}

Concerning the regression problem, we recall that a stochastic maximum-flow problem is a problem where the edge capacities in a flow network are modeled as random variables and the target is to find the distribution of the maximum flow (e.g., see \cite{DING20152056}). 
The task is to approximate with a NN model the function
\begin{equation}\label{eq:flowfunc}
\begin{aligned}
  \Phi \colon \R_+^n & \longrightarrow \R_+^m \\
  \v{c} & \longmapsto \Phi(\v{c}) = \v{\varphi}
\end{aligned}
\end{equation}
where $\v{c}:=(c_1,\ldots ,c_n)\in\R^n_+$ is the vector of the \emph{capacities} of all the $n$ edges of the network and $\v{\varphi}:=(\varphi_1, \ldots ,\varphi_m)\in\R^m_+$ is the flow vector corresponding to the $m$ incoming edges of the network's sink that generate the maximum flow; in other words, the maximum flow corresponding to $\v{c}$ is $\varphi:=\norm{\Phi(\v{c})}_1=\sum_{j=i}^m \varphi_j$.

To address this regression task, we build the GINNs and the EWGINNs with respect to the adjacency matrix of the \emph{line graph} of the flow network; i.e., on the graph where the vertices correspond to edges of the network and two vertices are connected if the corresponding edges in the network share at least one vertex. 
We refer to \cite{GINN} for more details about the formulation of this RoGN task for learning the maximum flow of a stochastic flow network (SFN).

\subsection{Performance Measures}

Let $\wh{\Phi}$ denote a NN model trained for learning \eqref{eq:flowfunc} and let $\mathcal{P}$ be a test set used for measuring the performances of the model. Then, denoted by $\wh{\v{\varphi}}:=\wh{\Phi}(\v{c})\in\R^m_+$, we define the following performance measures:
\begin{itemize}
    \item Average Mean Relative Error (MRE) of sink's incoming flows, with respect to the max-flow:
    \begin{equation}\label{eq:mre_elwise}
            \mathrm{MRE}_{av}(\mathcal{P}):=\frac{1}{m}\sum_{j=1}^m\left( \frac{1}{|\mathcal{P}|}\sum_{(\v{c}, \v{\varphi})\in\mathcal{P}}\frac{|\varphi_j - \wh{\varphi}_j|}{\varphi}\right)\,.
    \end{equation}
    This error measure describes the average quality of the NN in predicting the single elements $\varphi_1,\ldots ,\varphi_m$.
    \item Average max-flow MRE:
    \begin{equation}\label{eq:mre_tot}
            \mathrm{MRE}_{\varphi}(\mathcal{P}):\frac{1}{|\mathcal{P}|}\sum_{(\v{c}, \v{\varphi})\in\mathcal{P}}\frac{|\varphi - \wh{\varphi}|}{\varphi}
    \end{equation}
    This error measure describes the NN capability to predict the vector of fluxes $\wh{\v{\varphi}}$ such that the corresponding maxflow $\wh{\varphi}$ approximates the true maxflow $\varphi$.
\end{itemize}

\subsection{Data, Model Architectures, and Hyperparameters}

We run our experiments on the same data reported in \cite{GINN} for two randomly generated SFNs: a network based on a Barab\'asi-Albert (BA) graph and a network based on an Erdos-R\'enyi (ER) graph. 
Each of the datasets $\mathcal{D}_{\rm BA}$ and $\mathcal{D}_{\rm ER}$ consists of $10\, 000$ samples of capacity vectors and corresponding flow vectors. 

In this work, we focus on the harder case illustrated in \cite{GINN}: for each SFN, we train the \red{EWGINN and GINN models on 500 samples ($20\%$ used as validation set), measuring the errors MRE$_{av}$ and MRE$_\varphi$ on a test set of 3000 samples. 
Then, we compare the performances obtained by EWGINNs and GINNs, looking at the ``median models'' of each training configuration, where the median is computed with respect to the five initializations generated through the different random seeds.}

For a fair comparison, the architectures and hyperparameters of the EWGINN \red{and GINN models are the same and follow the criteria indicated in \cite{GINN}. 
Specifically, we build 60 EWGINN and GINN models configurations, respectively,} for each SFN, varying among these parameters: hidden layers' activation function $\sigma \in  \{\textit{ELU}, \textit{swish}, \textit{softplus}\}$, depth $H\in\{3,5,7,9\}$ for $\mathcal{G}_{\rm BA}$ and $H\in\{4,9,14,19\}$ for $\mathcal{G}_{\rm ER}$, output features of each \red{EWGI/GI} layer $F\in\{1,5,10\}$, output layer's pooling operation (if $F>1$) $\textit{pool}\in \{\textit{reduce\_max}, \textit{reduce\_mean}\}$. 
Also, the training options are the same used in \cite{GINN}: Adam optimizer \cite{Kingma2015_ADAM} (learning rate $\epsilon$=0.002, moment decay rates $\beta_1=0.9, \beta_2=0.999$), \red{early stopping regularization \cite{Goodfellow-et-al-2016} (550 epochs of patience, starting epoch 200, restore best weights), reduction on plateau for the learning rate (reduction factor $\alpha=0.5$, 50 epochs of patience, minimum $\epsilon=10^{-6}$). Each model configuration is trained five times, with respect to five different random seeds, respectively, for a total number of $1200$ trained models ($600$ per SFN).}

\subsection{Analysis of the Results}

\begin{figure}[htb!]
    \centering
    \includegraphics[trim=1.9cm .5cm 2.4cm 1.5cm,clip,width=0.45\textwidth]{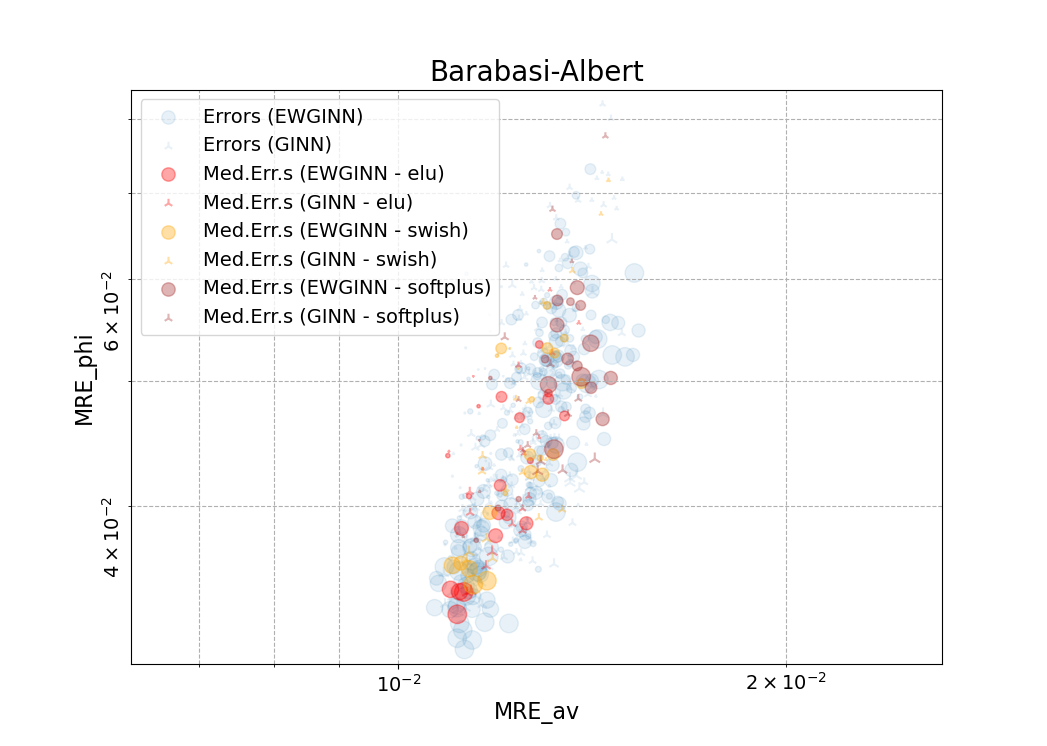}
    \caption{$\mathcal{G}_{\rm BA}$. Performances of GINN and EWGINN models in the $(\mathrm{MRE}_{av}, \mathrm{MRE}_\varphi)$ plane. Marker sizes are proportional to the number of NN weights.}
    \label{fig:BAres}
\end{figure}

\begin{figure}[htb!]
    \centering
    \includegraphics[trim=1.9cm .5cm 2.4cm 1.5cm,clip,width=0.45\textwidth]{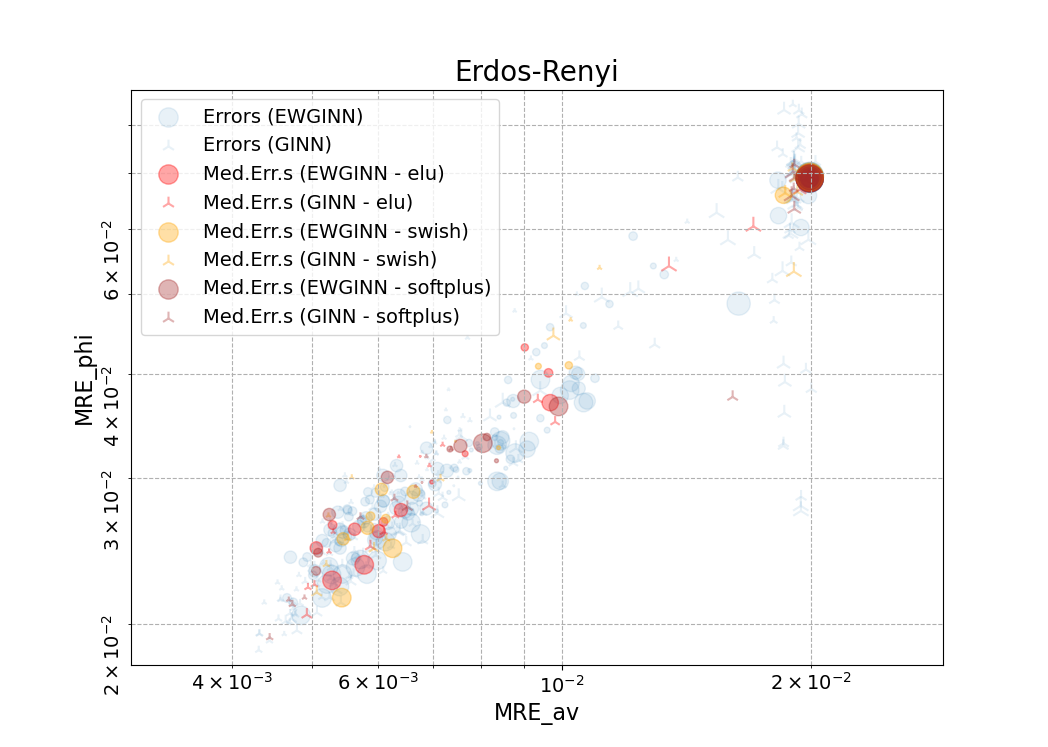}
    \caption{$\mathcal{G}_{\rm ER}$. Performances of GINN and EWGINN models in the $(\mathrm{MRE}_{av}, \mathrm{MRE}_\varphi)$ plane. Marker sizes are proportional to the number of NN weights.}
    \label{fig:ERres}
\end{figure}

Figures \ref{fig:BAres} and \ref{fig:ERres} compare the errors between classical GINNs (\red{tripod markers}) and EWGINN (\red{circular markers}). The error plane shows the MRE$_{av}$ error on the x-axis and the MRE$_\varphi$ on the y-axis. \red{The dot sizes are proportional to the number of NN weights, and dots corresponding to ``median models'' are colored according to the activation functions. Each median model is computed, among the five random seeds for each configuration, with respect to the distance of (MRE$_{av}$, MRE$_{\varphi}$) from the origin of the plane.}

\begin{figure}[htb!]
    \centering
    \includegraphics[trim=1.cm .3cm 2.cm 1.25cm,clip,width=0.375\textwidth]{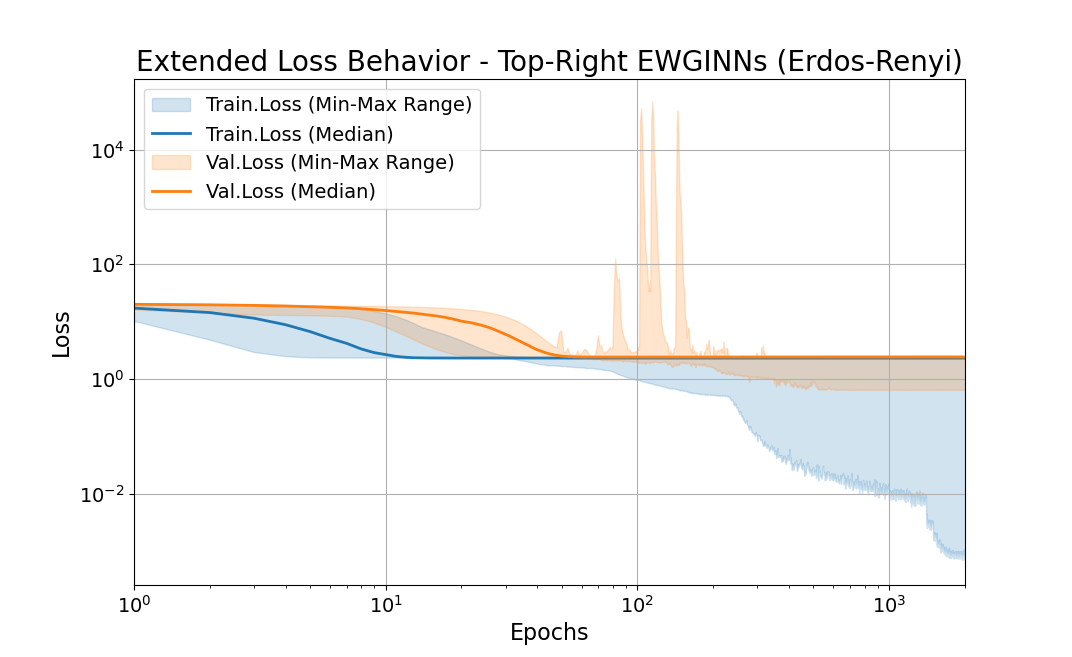}
    \caption{Training and validation loss of the EWGINN corresponding to the top-rightmost dots in \Cref{fig:ERres}.}
    \label{fig:badER_loss}
\end{figure}

\red{
We observe that the performance of GINNs and EWGINNs are comparable both in $\mathcal{G}_{\rm BA}$ and $\mathcal{G}_{\rm ER}$, but different behaviors characterize them.
In $\mathcal{G}_{\rm BA}$ (\Cref{fig:BAres}), the comparison is almost straightforward: GINNS and EWGINNs show similar trends in their performance, varying configurations and random seeds; however, EWGINNs show general better performance than GINNs for this SFN.

On the other hand, in $\mathcal{G}_{\rm ER}$ (\Cref{fig:ERres}), the performance trends of GINNs and EWGINNs are different. In particular, we observe that the EWGINN performances appear ``more stable'' than the GINN ones, varying configurations and random seeds. Indeed, we observe that EWGINN error points present a rather compact distribution, showing good regularization abilities of EWGINNs on the RoGN task (i.e., they reduce equally both the errors); on the contrary, GINN errors exhibit a sparser distribution; then, these models sometimes learn the task focusing more on MRE$_{\varphi}$ than MRE$_{av}$. We point out that the ability to learn the RoGN task without preferences in reducing one of the two errors is well appreciated. Indeed, as observed in \cite{GINN}, a small MRE$_{\varphi}$ and large MRE$_{av}$ can be the result of symmetric underestimation/overestimation of the single flows. Therefore, even if the best performances are reached by a subset of GINNs, the EWGINNs prove to be more reliable, varying hyperparameters and initializations, while maintaining very good performances.
These observations in $\mathcal{G}_{\rm ER}$ can be explained by the more chaotic connection structure of the SFN, if compared to $\mathcal{G}_{\rm BA}$; Indeed, EWGINNs has a clear advantage in regularizing their training, thanks to the property of rescaling the incoming information of nodes through additional weights.

We conclude by observing a cluster of GINNs and EWGINNs with poor performances for $\mathcal{G}_{\rm ER}$, constrained in an extremely small region (top-right corner, \Cref{fig:ERres}); the reason is an issue with early stopping. Specifically, a relatively fast reduction of the validation loss (VL) happens, resulting in a temporary overfitting or non-decreasing-VL phenomenon. This induces an interruption of the training due to the early stopping. 
Nonetheless, by removing the early stopping and increasing the training epochs, we observe that the overfitting phenomenon tends to disappear (see \Cref{fig:badER_loss}); moreover, in some cases, the VL starts to decrease again after some epochs.
Therefore, we conclude that the larger representational capacity of EWGINNs is an advantage but requires more careful tuning of the training hyperparameters. On the contrary, the GINNs are less influenced by this behavior because of their reduced size; nonetheless, when they ``escape'' from such situations, they usually fall into the problem of focusing more on MRE$_{\varphi}$ than MRE$_{av}$.

We defer to future work an in-depth analysis of EWGINNs by varying the training hyperparameters, such as the early stopping patience. 
}

\section{Conclusion}\label{sec:conclusions}

In this work, we proposed a novel type of GI layer: the Edge-Wise GI layer. 
Compared with the original GI layers, each node of an EWGI layer is equipped with an additional weight for rescaling the incoming message. 
This enables improved representational capacity and breaks the symmetry of GI layers, where nodes with the same neighborhood invariably receive the same message from the previous layer.
To analyze the performance of the newly proposed layers, we compared EWGINNs and GINNs on two benchmark RoGN tasks based on two SFNs\red{, respectively: one with graph connectivity concentrated on a few more central nodes ($\mathcal{G}_{\rm BA}$); one characterized by a random structure ($\mathcal{G}_{\rm ER}$).

The numerical experiments show comparable performance between GINNs and EWGINNs on both SFNs, though we observe distinct behaviors. EWGINNs perform better than GINNs on $\mathcal{G}_{\rm BA}$; on the other hand they exhibit improved regularization abilities on $\mathcal{G}_{\rm ER}$, maintaining comparable performance with GINNs. These results highlight the advantages of EWGINNs, particularly in handling the chaotic structure of $\mathcal{G}_{\rm ER}$, though their larger representational capacity demands more careful hyperparameter tuning. Observations of poor performance models caused by a too-early stopping suggest future studies focused on optimizing training configurations for EWGINNs. Future work will focus on applications to real-world problems.
}

\section*{Acknowledgements}

F.D., S.P., and F.V. acknowledge that this study was carried out within the FAIR-Future Artificial Intelligence Research and received funding from the European Union Next-GenerationEU (PIANO NAZIONALE DI RIPRESA E RESILIENZA (PNRR)–MISSIONE 4 COMPONENTE 2, INVESTIMENTO 1.3---D.D. 1555 11/10/2022, PE00000013). 
A.M. acknowledges support from the FINDHR project that received funding from the European Union’s Horizon Europe research and innovation program under grant agreement No. 101070212.
This manuscript reflects only the authors’ views and opinions; neither the European Union nor the European Commission can be considered responsible for them.
F.D. and S.P. acknowledge support from Italian MUR PRIN project 20227K44ME, Full and Reduced order modeling of coupled systems: focus on non-matching methods and automatic learning (FaReX).

\noindent\emph{Code Availability:}
The code for implementing the EWGI layers introduced in this paper is available at: \url{https://github.com/Fra0013To/GINN/tree/ewginn_dev}.





 \bibliographystyle{elsarticle-num} 
 \bibliography{DellaPapers,aiPapers,otherPapers,NetworksPapers,GNN_papers}





\end{document}